\documentclass{article}
\usepackage{ijcai23}

\usepackage{times}
\usepackage{soul}
\usepackage{url}
\usepackage[hidelinks]{hyperref}
\usepackage[utf8]{inputenc}
\usepackage[small]{caption}
\usepackage{graphicx}
\usepackage{amsmath,amsfonts}
\usepackage{amsthm}
\usepackage{booktabs}
\usepackage{algorithm}
\usepackage{algorithmic}
\usepackage{subfigure,multirow}
\usepackage{color}
\usepackage{booktabs}
\usepackage{arydshln}
\usepackage{hyperref,MnSymbol}
\usepackage[switch]{lineno}

\title{MM-PCQA: Multi-Modal Learning for No-reference Point Cloud Quality Assessment}

\author{
Zicheng Zhang$^1$
\and
Wei Sun$^1$\and
Xiongkuo Min$^1$\and
Qiyuan Wang$^3$\and
Jun He$^3$ \and
Quan Zhou$^3$ \And
Guangtao Zhai$^{1,2}$
\affiliations
$^1$Institute of Image Communication and Network Engineering, Shanghai Jiao Tong University\\
$^2$MoE Key Lab of Artificial Intelligence, AI Institute, Shanghai Jiao Tong University\\
$^3$Bilibili Inc, Shanghai, China
\emails
zzc1998@sjtu.edu.cn
}

\begin{document}

\maketitle

\begin{abstract}
    The visual quality of point clouds has been greatly emphasized since the ever-increasing 3D vision applications are expected to provide cost-effective and high-quality experiences for users.  Looking back on the development of point cloud quality assessment (PCQA), the visual quality is usually evaluated by utilizing single-modal information, i.e., either extracted from the 2D projections or 3D point cloud. The 2D projections contain rich texture and semantic information but are highly dependent on viewpoints, while the 3D point clouds are more sensitive to geometry distortions and invariant to viewpoints. Therefore, to leverage the advantages of both point cloud and projected image modalities, we propose a novel no-reference \underline{M}ulti-\underline{M}odal \underline{P}oint \underline{C}loud \underline{Q}uality \underline{A}ssessment (\textbf{MM-PCQA}) metric. In specific, we split the point clouds into sub-models to represent local geometry distortions such as point shift and down-sampling. Then we render the point clouds into 2D image projections for texture feature extraction. To achieve the goals, the sub-models and projected images are encoded with point-based and image-based neural networks. Finally, symmetric cross-modal attention is employed to fuse multi-modal quality-aware information. Experimental results show that our approach outperforms all compared state-of-the-art methods and is far ahead of previous no-reference PCQA methods, which highlights the effectiveness of the proposed method. The code is available at https://github.com/zzc-1998/MM-PCQA.
\end{abstract}

\section{Introduction}
Point cloud has been widely adopted in practical applications such as virtual/augmented reality \cite{park2008multiple}, automatic driving \cite{cui2021deep}, and video post-production \cite{mekuria2016design} due to its ability of representing the 3D world. Consequently, plenty of works have been carried out to deal with point cloud classification \cite{grilli2017review,ku2018joint,wang2019frustum,vora2020pointpainting,xie2020pi,yoo20203d,chen2020object}, detection \cite{cui2021deep}, and segmentation \cite{cheng2021sspc,liutoposeg}. However, point cloud quality assessment (PCQA) has gained less attention. PCQA aims to predict the visual quality level of point clouds, which is vital for providing useful guidelines for simplification operations and compression algorithms in applications such as metaverse and virtual/augmented reality (VR/AR) \cite{fan2022d} to not negatively impact users' quality of experience (QoE). Moreover, the point clouds representing vivid objects/humans are usually more complex in geometric structure and contain large amounts of dense points along with color attributes, which makes the PCQA problem very challenging.

\begin{figure}[t]
    \centering
    \includegraphics[width=0.75\linewidth]{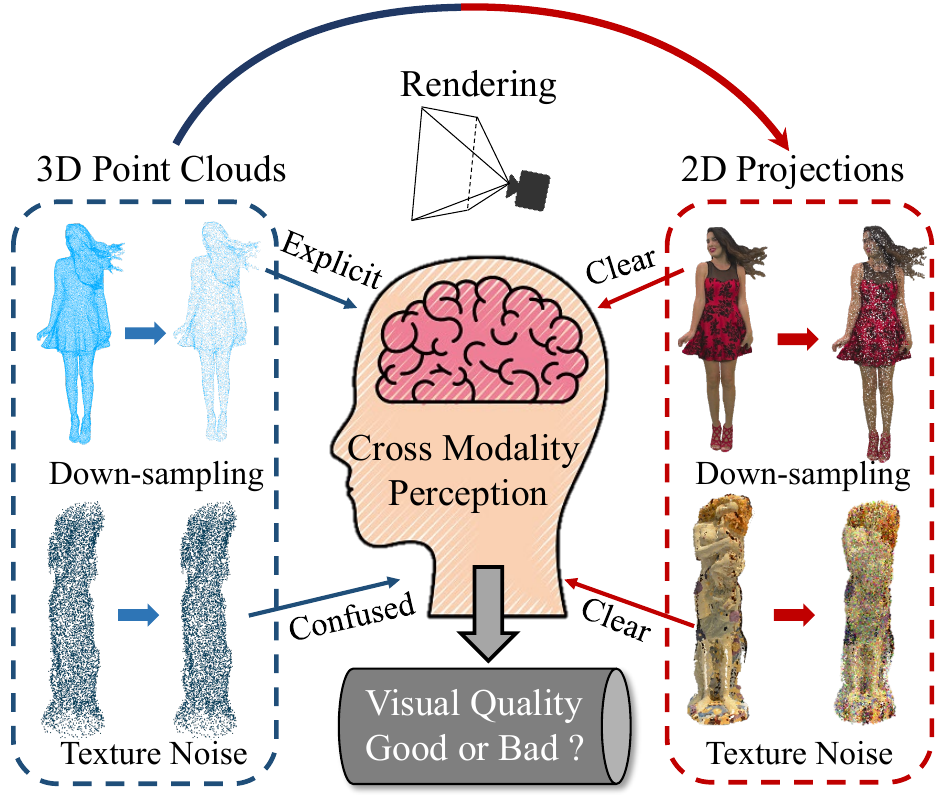}
    \caption{Examples of reflected distortions. The point clouds can explicitly reveal the geometry down-sampling distortion while failing to recognize texture noise unless the projections are involved, which raises the need for multi-modal perception. }
    \label{fig:introduction}
    \vspace{-0.5cm}
\end{figure}

Generally speaking, PCQA methods can be divided into full-reference PCQA (FR-PCQA), reduced-reference (RR-PCQA), and no-reference PCQA (NR-PCQA) methods according to the involvement extent of the reference point clouds. However, the pristine reference point clouds are not always available in many practical situations, thus NR-PCQA has a wider range of applications. Hence, we focus on the NR-PCQA in this paper. Reviewing the development of NR-PCQA, most metrics are either based on the point cloud features extracted by statistical models \cite{zhang2021mesh,zhang2021no} and end-to-end neural networks \cite{liu2022point} or based on the projected image features obtained via hand-crafted manners \cite{meynet2020pcqm,yang2020graphsim,alexiou2020pointssim} or 2D convolutional neural networks (CNN) \cite{liu2021pqa,fan2022no,zhang2022treating}. Such methods fail to jointly make use of the information from 3D point clouds along with 2D projections, thus resulting in unsatisfactory performance.


Therefore, to boost the performance of PCQA, we propose a multi-modal learning strategy for NR-PCQA, which extracts quality-aware features not only from the 3D point clouds but also from the 2D projections. There are two main reasons to adopt this strategy. First, point clouds can be perceived in both 2D/3D scenarios. We can view point clouds from 2D perspective via projecting them on the screen or directly watch point clouds in 3D format using the VR equipment. Thus multi-modal learning is able to cover more range of practical situations. Second, different types of distortions have divergent visual influence on different modalities. As shown in Fig. \ref{fig:introduction}, the structural damage and geometry down-sampling are more obvious in the point cloud modality while the image modality is more sensitive to texture distortions caused by color quantization and color noise. Moreover, it is easier to extract quality-aware semantic information from the image modality. Thus, the proposed multi-modal learning fashion can make up for the deficiencies and take advantage of both modalities.
 Further, considering that the local patterns such as smoothness and roughness are very important for quality assessment, we first propose to split the point cloud into sub-models rather than sampling points for analysis, which has been previously adopted for extracting the low-level pattern features of the point cloud \cite{chetouani2021deep,you2021patch}. The image projections are acquired by rendering the point clouds from several viewpoints with a fixed viewing distance to maintain the consistency of the texture scale. Then a point cloud encoder and an image encoder are used to encode the point clouds and projected images into quality-aware embeddings respectively, which are subsequently strengthened by symmetric cross-modality attention. Finally, the quality-aware embeddings are decoded into quality values with fully-connected layers. The main contributions of this paper are summarized as follows:

\begin{itemize}
    \item We propose a no-reference \underline{M}ulti-\underline{M}odal \underline{P}oint \underline{C}loud \underline{Q}uality \underline{A}ssessment (\textbf{MM-PCQA}) to interactively use the information from both the point cloud and image modalities. \textbf{To the best of our knowledge, we are the first to introduce multi-modal learning into the PCQA field.}
    \item To preserve the local patterns that are vital for visual quality, we propose to split the point cloud into sub-models rather than sampling points as the input of the point cloud encoder. To better incorporate the multi-modal features, we employ cross-modal attention to model the mutual relationship between the quality-aware features extracted from two modalities.
    \item Extensive experiments show that MM-PCQA achieves the best performance among the compared state-of-the-art methods (even including the FR-PCQA methods). The ablation studies reveal the contributions of different modalities, the patch-up strategy, and cross-modal attention, demonstrating the effectiveness of the proposed framework.
\end{itemize}

\begin{figure*}
    \centering
    \includegraphics[width = 0.98\linewidth]{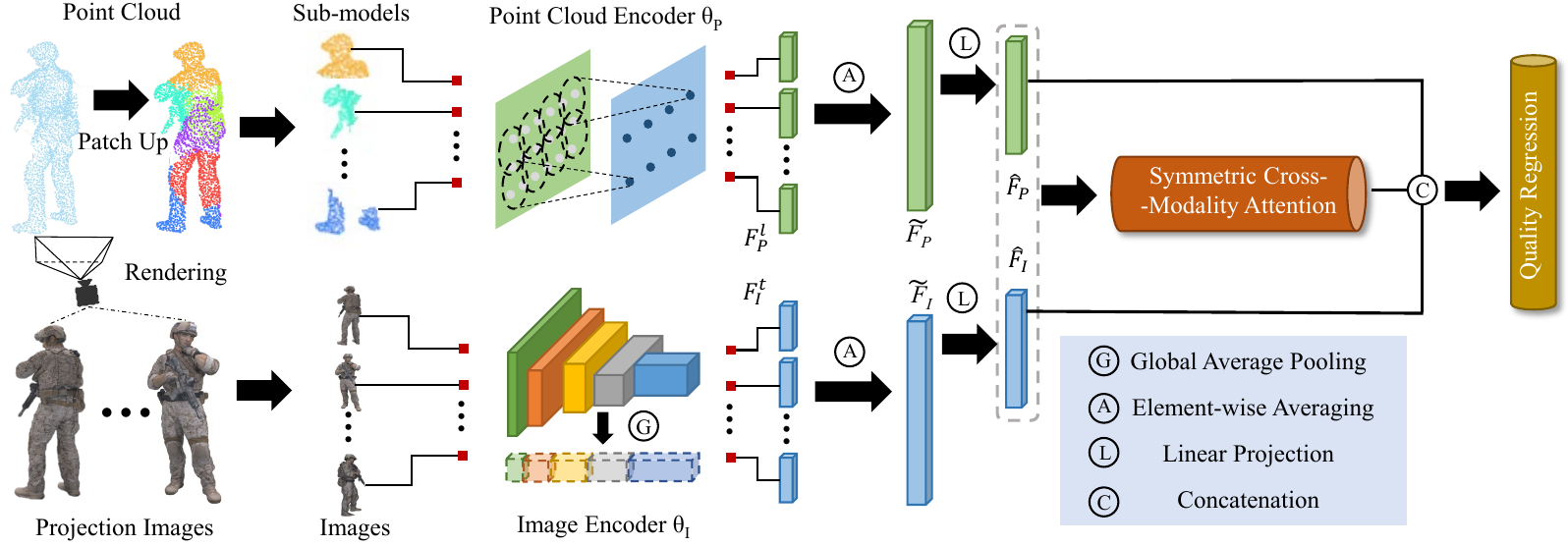}
    \caption{The framework of the proposed method.}
    \label{fig:framework}
\end{figure*}

\section{Related Work}
\subsection{Quality Assessment for Point Cloud}
In the early years of PCQA development, some simple point-based FR-PCQA methods are proposed by MPEG, such as p2point \cite{mekuria2016evaluation} and p2plane \cite{tian2017geometric}. To further deal with colored point clouds, point-based PSNR-yuv \cite{torlig2018novel} is carried out. Since the point-level difference is difficult to reflect complex distortions, many well-designed FR-PCQA metrics are proposed to employ structural features and have achieved considerable performance, which includes PCQM \cite{meynet2020pcqm}, GraphSIM \cite{yang2020graphsim}, PointSSIM \cite{alexiou2020pointssim}, etc. 

To cover more range of practical applications and inspired by NR image quality assessment \cite{zhang2022nor,zhang2022dual,zhang2021ano}, some NR-PCQA methods have been proposed as well.  Chetouani $et$  $al.$ \cite{chetouani2021deep} extract patch-wise hand-crafted features and use classical CNN models for quality regression. PQA-net \cite{liu2021pqa} utilizes multi-view projection for feature extraction. Zhang $et$  $al.$ \cite{zhang2021no} use several statistical distributions to estimate quality-aware parameters from the geometry and color attributes' distributions. Fan $et$  $al.$ \cite{fan2022no} infer the visual quality of point clouds via the captured video sequences. Liu $et$  $al.$ \cite{liu2022point} employ an end-to-end sparse CNN for quality prediction. Yang $et$  $al.$ \cite{yang2022no} further transfer the quality information from natural images to help understand the point cloud rendering images' quality via domain adaptation. The mentioned methods are all based on single-modal information, thus failing to jointly incorporate the multi-modal quality information.

\subsection{Multi-modal Learning for Point Cloud}
Many works \cite{radford2021learning,cheng2020look} have proven that multi-modal learning is capable of strengthening feature representation by actively relating the compositional information across different modalities such as image, text, and audio. Afterwards, various works utilize both point clouds and image data to improve the understanding of 3D vision, which are mostly targeted at 3D detection. Namely, AVOD \cite{ku2018joint} and MV3D \cite{chen2020object} make region proposals by mapping the LiDAR point clouds to bird's eye view. Then Qi $et$  $al.$ \cite{qi2018frustum} and Wang $et$  $al.$ \cite{wang2019frustum} attempt to localize the points by proposing 2D regions with 2D CNN object detector and then transforming the corresponding 2D pixels to 3D space.  Pointpainting \cite{vora2020pointpainting} projects the points into the image segmentation mask and detects 3D objects via LiDAR-based detector. Similarly, PI-RCNN \cite{xie2020pi} employs semantic feature maps and makes predictions through point-wise fusion. 3D-CVF  \cite{yoo20203d} transfers the camera-view features using auto-calibrated projection and fuses the multi-modal features with gated fusion networks. More recently, some works \cite{afham2022crosspoint,zhang2022cat} try to make use of multi-modal information in a self-supervised manner and transfer the learned representation to the downstream tasks.

\begin{figure}[!ht]
    \centering
    \subfigure[Ping-pong bat]{\begin{minipage}[t]{0.31\linewidth}
                \centering
                \includegraphics[width = 2cm]{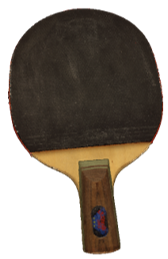}
                \end{minipage}}
    \subfigure[Anchors]{\begin{minipage}[t]{0.31\linewidth}
                \centering
                \includegraphics[width = 2.16cm]{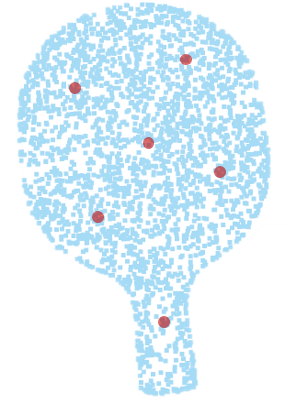}
                \end{minipage}}
    \subfigure[Sub-models]{\begin{minipage}[t]{0.31\linewidth}
                \centering
                \includegraphics[width = 2.32cm]{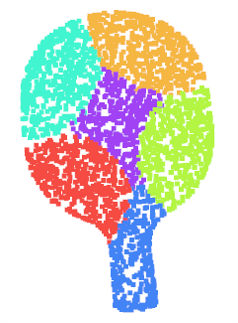}
                \end{minipage}}
    \subfigure[House]{\begin{minipage}[t]{0.31\linewidth}
                \centering
                \includegraphics[width = 2cm]{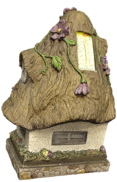}
                \end{minipage}}
    \subfigure[Anchors]{\begin{minipage}[t]{0.31\linewidth}
                \centering
                \includegraphics[width = 2.16cm]{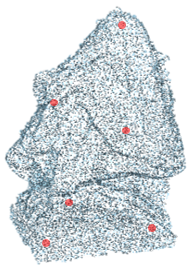}
                \end{minipage}}
    \subfigure[Sub-models]{\begin{minipage}[t]{0.31\linewidth}
                \centering
                \includegraphics[width = 2.32cm,height=3.2cm]{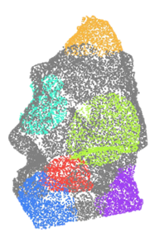}
                \end{minipage}}
    \subfigure[Flowerpot]{\begin{minipage}[t]{0.31\linewidth}
                \centering
                \includegraphics[width = 2.16cm]{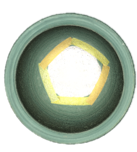}
                \end{minipage}}
    \subfigure[Anchors]{\begin{minipage}[t]{0.31\linewidth}
                \centering
                \includegraphics[width = 2.16cm]{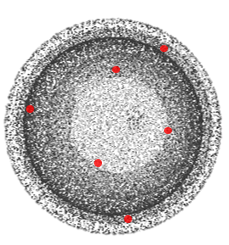}
                \end{minipage}}
    \subfigure[Sub-models]{\begin{minipage}[t]{0.31\linewidth}
                \centering
                \includegraphics[width = 2.32cm]{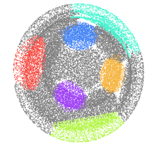}
                \end{minipage}}
    \caption{Examples of the patch-up process with $N_{\delta}=6$. (a), (d), and (g) represent the source point clouds. (b), (e), and (h) describe the 6 anchor points after the farthest point sampling. (c), (f), and (i) exhibit the patch-up results after KNN operation, which generates 6 sub-models. The ping-pong bat point cloud contains fewer points, therefore nearly all the points can be covered in the 6 sub-models. However, not all points in the house and flowerpot point clouds can be include in the 6 sub-models. The gray part shown in (f) and (i) represents the points that are not covered in the sub-models.}
    \label{fig:patch}
\end{figure}

\section{Proposed Method}
The framework overview is clearly exhibited in Fig. \ref{fig:framework}. The point clouds are first segmented into sub-models and put into the point cloud encoder $\theta_{P}$. The projected images are directly rendered from the colored point clouds and put into the image encoder $\theta_{I}$. Subsequently, the quality-aware encoder features are optimized with the assistance of symmetric cross-modality attention. Finally, the features are concatenated and decoded into the quality values via the quality regression. 

\subsection{Preliminaries} Suppose we have a colored point cloud $\mathbf{P} = \{g_{(i)} , c_{(i)} \}_{i=1}^{N}$, where $g_{(i)} \in \mathbb{R}^{1 \times 3} $ indicates the geometry coordinates, $c_{(i)} \in \mathbb{R}^{1 \times 3} $ represents the attached RGB color information, and $N$ stands for the number of points.  The point cloud modality $\mathbf{\hat{P}}$ is obtained by normalizing the original geometry coordinates and the image modality $\mathbf{I}$ is generated by rendering the colored point cloud $\mathbf{P}$ into 2D projections. Note that $\mathbf{\hat{P}}$ contains no color information.

\subsection{Point Cloud Feature Extraction}
\label{sec:pc}
It is natural to transfer mainstream 3D object detectors such as PointNet++ \cite{qi2017pointnet++} and DGCNN \cite{wang2019dynamic} to visual quality representation of point clouds. However, different from the point clouds used for classification and segmentation, the high-quality point clouds usually contain large numbers of dense points, which makes it difficult to extract features directly from the source points unless utilizing down-sampling. Nevertheless, the common sampling methods are aimed at preserving semantic information whereas inevitably damaging the geometry patterns that are crucial for quality assessment. To avoid the geometry error caused by the down-sampling and preserve the smoothness and roughness of local patterns, we propose to gather several local sub-models from the point cloud for geometric structure feature representation. 

Specifically, given a normalized point cloud $\mathbf{\hat{P}}$, we employ the farthest point sampling (FPS) to obtain $N_{\delta}$ anchor points $\{\delta_{m}\}_{m=1}^{N_{\delta}}$. For each anchor point, we utilize K nearest neighbor (KNN) algorithm to find $N_{s}$ neighboring points around the anchor point and form such points into a sub-model:

\begin{equation}
   {\rm S} = \{\mathop{\boldsymbol{\rm{KNN}}}\limits_{k=N_{s}}(\delta_{m})\}_{m=1}^{N_{\delta}},
\end{equation}
where $\rm{S}$ is the set of sub-models, $\boldsymbol{\rm{KNN}}(\cdot)$ denotes the KNN operation and $\delta_{m}$ indicates the $m$-th farthest sampling point. An illustration of the patch-up process is exhibited in Fig. \ref{fig:patch}. It can be seen that the local sub-models are capable to preserve the local patterns.
Then a point cloud feature encoder $\theta_{P}(\cdot)$ is employed to map the obtained sub-models to quality-aware embedding space:

\begin{equation}
\begin{aligned}
   \rm{F_{P}} &= \{\theta_{P}(S_{l})\}_{l=1}^{N_{\delta}}, \\
   \widetilde{F}_{P} &= \frac{1}{N_{\delta}}\sum_{l = 1}^{N_{\delta}}F_{P}^{l},
\end{aligned} 
\end{equation}
where $F_{P}^{l} \in \mathbb{R}^{1 \times C_{P}}$ indicates the quality-aware embedding for the $l$-th sub-model $S_{l}$, $C_{P}$ represents the number of output channels of the point cloud encoder $\theta_{P}(\cdot)$, and $\widetilde{F}_{P}  \in \mathbb{R}^{1 \times C_{P}}$ is the pooled results after average fusion.

\begin{figure}[!t]
    \centering
    \includegraphics[width = 0.85\linewidth]{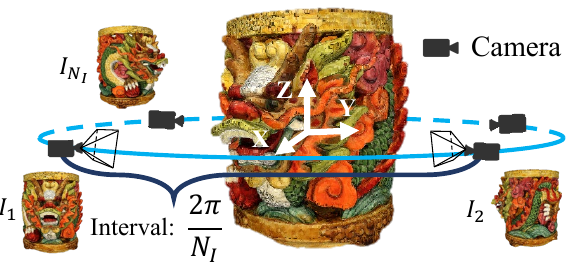}
    \caption{Illustration of the projections rendering process. Camera is rotated along the circular pathway with intervals of $\frac{2\pi}{N_{I}}$ to capture $N_{I}$ projections. }
    \label{fig:image}
    \vspace{-0.175cm}
\end{figure}

\subsection{Image Feature Extraction}
\label{sec:image}
$N_{I}$ image projections are rendered from the distorted colored point clouds from a defined circular pathway with a fixed viewing distance to keep the texture consistent as shown in Fig. \ref{fig:image}:

\begin{equation}
\left\{
\begin{aligned}
X^2 + Y^2 &= R^2, \\
Z &= 0, 
\end{aligned}
\right.
\end{equation}
where the pathway is centered on the point cloud's geometry center, $R$ indicates the fixed viewing distance and the $N_{I}$ image projections are captured with intervals of $\frac{2\pi}{N_{I}}$. The projections are rendered with the assistance of Open3D \cite{zhou2018open3d}.
Then we embed the rendered 2D images into quality-aware space with the 2D image encoder:

\begin{equation}
\begin{aligned}
   F_{I} &= \{\theta_{I}(I_{t})\}_{t=1}^{N_{I}}, \\
   \widetilde{F}_{I} &= \frac{1}{N_{I}}\sum_{t = 1}^{N_{I}}F_{I}^{t},
\end{aligned} 
\end{equation}
where $F_{I}^{t} \in \mathbb{R}^{1 \times C_{I}}$ denotes the quality-aware embedding for the $t$-th image projection $I_{t}$, $C_{I}$ represents the number of output channels of the 2D image encoder $\theta_{I}(\cdot)$, and $\widetilde{F}_{I}  \in \mathbb{R}^{1 \times C_{I}}$ is the pooled results after average fusion.

\begin{figure}[t]
    \centering
    \includegraphics[width=0.73\linewidth]{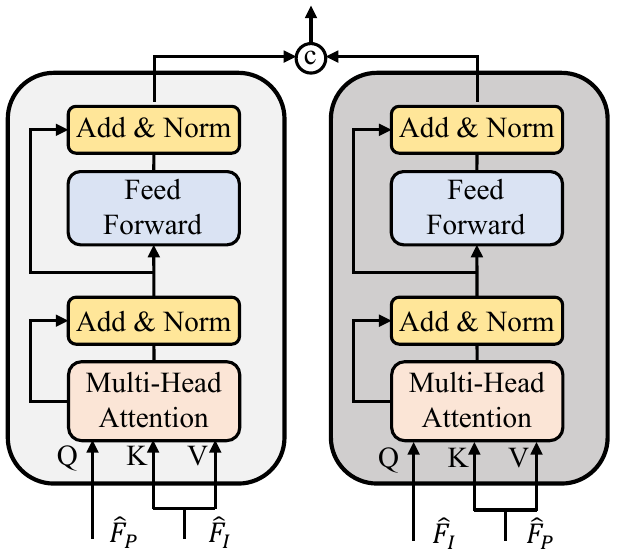}
    \caption{Illustration of the symmetric cross-modality attention module (SCMA). The point cloud embedding $\hat{F}_{P}$ is used to guide the attention learning of image embedding and the image embedding $\hat{F}_{I}$ is used to guide the attention learning of point cloud embedding respectively.}
    \label{fig:cross_modal}
\end{figure}

\subsection{Symmetric Cross-Modality Attention}
As shown in Fig \ref{fig:introduction}, the single modality may be incomplete to cover sufficient information for quality assessment, thus we propose a symmetric attention transformer block to investigate the interaction between the visual quality features gathered from the point cloud and image modalities. Given the intra-modal features $\widetilde{F}_{P}  \in \mathbb{R}^{1 \times C_{P}}$ and $\widetilde{F}_{I}  \in \mathbb{R}^{1 \times C_{I}}$ from the point clouds and images respectively, we adjust them to the same dimension with linear projection:
\begin{equation}
     \hat{F}_{P} = W_{P}\widetilde{F}_{P}, \quad \hat{F}_{I} = W_{I}\widetilde{F}_{I},
\end{equation}
where $\hat{F}_{P} \in \mathbb{R}^{1 \times C'}$ and $\hat{F}_{I} \in \mathbb{R}^{1 \times C'} $ are adjusted features, $W_{P}$ and $W_{I}$ are learnable linear mappings, and $C'$ is the number of adjusted channels. To further explore the discriminate components among the modalities, the multi-head attention module is utilized:

\begin{equation}
\begin{aligned}
     & \Gamma (Q,K,V) = (h_{1} \oplus h_{2} \cdots \oplus h_{n}) W, \\
     & h_{\mu} = \beta(QW_{\mu}^{Q},KW_{\mu}^{K},VW_{\mu}^{V})|_{\mu=1}^n,\\
     &\beta(Q,K,V) = {\rm softmax}(QK^{T}/\sqrt{d})V, \\
\end{aligned} 
\end{equation}
where $\Gamma(\cdot)$ indicates the multi-head attention operation, $\beta(\cdot)$ represents the attention function, $h_{\mu}$ is the $\mu$-th head, and $W$, $W_{Q}$, $W_{K}$, $W_{V}$ are learnable linear mappings. As illustrated in Fig. \ref{fig:cross_modal}, both point-cloud-based and image-based quality-aware features participate in the attention learning of the other modality. The final quality embedding can be concatenated by the intra-modal features and the guided multi-modal features obtained by the symmetric cross-modality attention module:

\begin{equation}
    \hat{F}_{Q} = \hat{F}_{P} \oplus \hat{F}_{I} \oplus \Psi(\hat{F}_{P},\hat{F}_{I}),
\end{equation}
where $\oplus(\cdot)$ indicates the concatenation operation, $\Psi(\cdot)$ stands for the symmetric cross-modality attention operation, and $\hat{F}_{Q}$ represents the final quality-aware features ($\Psi(\hat{F}_{P},\hat{F}_{I}) \in \mathbb{R}^{1 \times 2C'}$, $\hat{F}_{Q} \in \mathbb{R}^{1 \times 4C'}$).

\subsection{Quality Regression \& Loss Function}
Following common practice, we simply use two-fold fully-connected layers to regress the quality features $\hat{F}_{Q}$ into predicted quality scores.
For the quality assessment tasks, we not only focus on the accuracy of the predicted quality values but also lay importance on the quality rankings. Therefore, the loss function employed in this paper includes two parts: Mean Squared Error (MSE) and rank error. The MSE is utilized to keep the predicted values close to the quality labels, which can be derived as:

\begin{equation}
    L_{MSE} = \frac{1}{n} \sum_{\eta=1}^{n} (q_{\eta}-q_{\eta}')^2,
\end{equation}
where $q_{\eta}$ is the predicted quality scores, $q_{\eta}'$ is the quality labels of the point cloud, and $n$ is the size of the mini-batch. The rank loss can better assist the model to distinguish the quality difference when the point clouds have close quality labels. To this end, we use the differentiable rank function described in \cite{sun2022deep} to approximate the rank loss:

\begin{equation}
\begin{aligned}
    L_{rank }^{i j}\!=\!\max\! &\left(0,\left|q_{i}-q_{j}\right|\!-\!e\left(q_{i}, q_{j}\right)\! \cdot\! \left({q}_{i}'-{q}_{j}'\right)\right), \\
    &e\left(q_{i}, q_{j}\right)=\left\{\begin{array}{r}
1, q_{i} \geq q_{j}, \\
-1, q_{i}<q_{j},
\end{array}\right.
\end{aligned}
\end{equation}
where $i$ and $j$ are the corresponding indexes for two point clouds in a mini-batch and the rank loss can be derived as:
\begin{equation}
    L_{r a n k}=\frac{1}{n^{2}} \sum_{i=1}^{n} \sum_{j=1}^{n} L_{r a n k}^{i j},
\end{equation}
Then the loss function can be calculated as the weighted sum of MSE loss and rank loss:
\begin{equation}
    Loss=\lambda_{1}L_{MSE}+\lambda_{2} L_{r a n k}
\end{equation}
where $\lambda_{1}$ and $\lambda_{2}$ are used to control the proportion of the MSE loss and the rank loss.

\section{Experiments}

\begin{table*}[th]\small
\centering
\renewcommand\tabcolsep{2.8pt} 
\begin{tabular}{c:c:l|cccc|cccc|cccc}
\toprule
\multirow{2}{*}{Type} &\multirow{2}{*}{Modal} &\multirow{2}{*}{Methods}  & \multicolumn{4}{c|}{SJTU-PCQA} & \multicolumn{4}{c|}{WPC} & \multicolumn{4}{c}{WPC2.0}\\ 
\cline{4-15}
        & & & SRCC$\uparrow$       & PLCC$\uparrow$      & KRCC$\uparrow$     & RMSE$\downarrow$     & SRCC$\uparrow$      & PLCC$\uparrow$      & KRCC$\uparrow$       & RMSE$\downarrow$  & SRCC$\uparrow$      & PLCC$\uparrow$      & KRCC$\uparrow$       & RMSE$\downarrow$ \\ \hline
\multirow{8}{*}{FR}
& P & MSE-p2po  & 0.7294 & 0.8123 & 0.5617 & 1.3613 & 0.4558 & 0.4852 & 0.3182 & 19.8943 & 0.4315 & 0.4626 & 0.3082 & 19.1605           \\
& P & HD-p2po  & 0.7157 & 0.7753 & 0.5447 & 1.4475 &0.2786 &0.3972&0.1943 &20.8990 &0.3587 & 0.4561 & 0.2641 & 18.8976           \\
& P & MSE-p2pl & 0.6277 & 0.5940 & 0.4825 & 2.2815 & 0.3281 & 0.2695 &0.2249 & 22.8226 & 0.4136 & 0.4104 & 0.2965 & 21.0400           \\
& P & HD-p2pl  & 0.6441   & 0.6874    & 0.4565    & 2.1255 & 0.2827 & 0.2753 &0.1696 &21.9893  & 0.4074 & 0.4402 & 0.3174 & 19.5154  \\
& P & PSNR-yuv  & 0.7950 & 0.8170 & 0.6196 & 1.3151 & 0.4493 & 0.5304 & 0.3198 & 19.3119 & 0.3732 & 0.3557 & 0.2277 & 20.1465\\
& P & PCQM     & {0.8644}   & {0.8853}    & \bf\textcolor{blue}{0.7086}     & {1.0862}     &\bf\textcolor{blue}{0.7434}    & \bf\textcolor{blue}{0.7499}   & \bf\textcolor{blue}{0.5601}   & 15.1639    &0.6825 & 0.6923 & 0.4929 & 15.6314            \\
    & P & GraphSIM  & \bf\textcolor{blue}{0.8783}    & \bf\textcolor{blue}{0.8449}    & {0.6947}   & \bf\textcolor{blue}{1.0321}  & 0.5831    & 0.6163    & 0.4194   & 17.1939 & 0.7405 & 0.7512 & \bf\textcolor{blue}{0.5533} & \bf\textcolor{blue}{14.9922}\\
& P & PointSSIM    & 0.6867  & 0.7136  & 0.4964 & 1.7001  & 0.4542    & 0.4667    & 0.3278   & 20.2733  & 0.4810 & 0.4705 & 0.2978 & 19.3917 \\ \hdashline
 \multirow{7}{*}{NR} 
&I &BRISQUE  & 0.3975    & 0.4214  & 0.2966 & 2.0937  & 0.2614    & 0.3155  & 0.2088 & 21.1736  & 0.0820  & 0.3353	& 0.0487 & 21.6679
\\
&I &NIQE  &0.1379 &0.2420 &0.1009 &2.2622  &0.1136 & 0.2225 &0.0953 &23.1415 & 0.1865 & 0.2925 & 0.1335 & 22.5146
 \\
&I &IL-NIQE  & 0.0837 & 0.1603 & 0.0594 &2.3378 &0.0913 &0.1422 & 0.0853 &24.0133 & 0.0911 & 0.1233 & 0.0714 & 23.9987\\
&P &ResSCNN  & 0.8600 & 0.8100 & - & - & - & - & - & - & \bf\textcolor{blue}{0.7500} & \bf\textcolor{blue}{0.7200} &-&- \\
&I &PQA-net  & 0.8372   & 0.8586    & 0.6304 & {1.0719}  & {0.7026}    & {0.7122}    & {0.4939}   & \bf\textcolor{blue}{15.0812} & 0.6191 & 0.6426 & 0.4606 & 16.9756    \\
&P &3D-NSS     & 0.7144 & 0.7382  & 0.5174 & 1.7686    & 0.6479    & 0.6514    & 0.4417   & 16.5716 & 0.5077 & 0.5699 & 0.3638 & 17.7219 \\ 
&P+I &\textbf{MM-PCQA}     & \bf\textcolor{red}{0.9103}   & \bf\textcolor{red}{0.9226}   & \bf\textcolor{red}{0.7838} & \bf\textcolor{red}{0.7716} &\bf\textcolor{red}{0.8414}    & \bf\textcolor{red}{0.8556}    & \bf\textcolor{red}{0.6513}  & \bf\textcolor{red}{12.3506} &\bf\textcolor{red}{0.8023}    & \bf\textcolor{red}{0.8024}    & \bf\textcolor{red}{0.6202}  & \bf\textcolor{red}{13.4289} \\

                      \bottomrule
\end{tabular}
\caption{Performance comparison with state-of-the-art approaches on the SJTU-PCQA, WPC, and WPC2.0 databases. ‘P’ and ‘I’ stand for the point cloud and image modalities respectively. Best in {\bf\textcolor{red}{red}} and second in {\bf\textcolor{blue}{blue}}. }
\label{tab:experiment}
\end{table*}


\subsection{Databases}
To test the performance of the proposed method, we employ the subjective point cloud assessment database (SJTU-PCQA) \cite{yang2020predicting}, the Waterloo point cloud assessment database (WPC) proposed by \cite{liu2022perceptual}, and the WPC2.0 database \cite{liu2021reduced} for validation. The SJTU-PCQA database includes 9 reference point clouds and each point cloud is corrupted with seven types of distortions (compression, color noise, geometric shift, down-sampling, and three distortion combinations) under six strengths, which generates 378 = 9$\times$7$\times$6 distorted point clouds in total.
The WPC database contains 20 reference point clouds and augments each point cloud with four types of distortions (down-sampling, Gaussian white noise, Geometry-based Point Cloud Compression (G-PCC), and Video-based Point Cloud Compression (V-PCC)), which generates 740 = 20$\times$37 distorted point clouds. The WPC2.0 databases provides 16 reference point clouds and degradaes the point clouds with 25 V-PCC settings, which generates 400 = 16$\times$25 distorted point clouds.

\subsection{Implementation Details}
The Adam optimizer \cite{kingma2014adam} is utilized with weight decay 1e-4, the initial learning rate is set as 5e-5, and the batch size is set as 8. The model is trained for 50 epochs by default. Specifically, We set the point cloud sub-model size $N_{s}$ as 2048, set the number of sub-models $N_{\delta} = 6$, and set the number of image projections $N_{I}=4$. The projected images with the resolution of 1920$\times$1080$\times$3 are randomly cropped into image patches at the resolution of 224$\times$224$\times$3 as the inputs (the white background is removed from the projected images). The PointNet++ \cite{qi2017pointnet++} is utilized as the point cloud encoder and the ResNet50 \cite{he2016deep} is used as the image encoder, where the ResNet50 is initialized with the pre-trained model on the ImageNet database \cite{deng2009imagenet}. The multi-head attention module employs 8 heads and the feed-forward dimension is set as 2048. The weights $\lambda_{1}$ and $\lambda_{2}$ for $L_{MSE}$ and $ L_{r a n k}$ are both set as 1. 

Following the practices in \cite{fan2022no}, the k-fold cross validation strategy is employed for the experiment to accurately estimate the performance of the proposed method. Since the SJTU-PCQA, WPC, and WPC2.0 contain 9, 20, 16 groups of point clouds respectively, 9-fold, 5-fold, and 4-fold cross validation is selected for the three database to keep the train-test ratio around 8:2. The average performance is recorded as the final results.
It's worth noting that there is no content overlap between the training and testing sets. We strictly retrain the available baselines with the same database split set up to keep the comparison fair. What's more, for the FR-PCQA methods that require no training, we simply validate them on the same testing sets and record the average performance.

\subsection{Competitors and Evaluation Criteria}
14 state-of-the-art quality assessment methods are selected for comparison, which consist of 8 FR-PCQA methods and 6 NR-PCQA methods. The FR-PCQA methods include MSE-p2point (MSE-p2po) \cite{mekuria2016evaluation}, Hausdorff-p2point (HD-p2po) \cite{mekuria2016evaluation}, MSE-p2plane (MSE-p2pl) \cite{tian2017geometric}, Hausdorff-p2plane (HD-p2pl) \cite{tian2017geometric}, PSNR-yuv \cite{torlig2018novel}, PCQM \cite{meynet2020pcqm}, GraphSIM \cite{yang2020graphsim}, and PointSSIM \cite{alexiou2020pointssim}.  The NR-PCQA methods include BRISQUE \cite{mittal2012brisque}, NIQE \cite{mittal2012making}, IL-NIQE \cite{zhang2015feature}, ResSCNN \cite{liu2022point}, PQA-net \cite{liu2021pqa}, and 3D-NSS \cite{zhang2021no}. Note that BRISQUE, NIQE, IL-NIQE are image-based quality assessment metrics and are validated on the same projected images.
Furthermore, to deal with the scale differences between the predicted quality scores and the quality labels, a five-parameter logistic function is applied to map the predicted scores to subjective ratings, as suggested by \cite{antkowiak2000final}.

Four mainstream evaluation criteria in the quality assessment field are utilized to compare the correlation between the predicted scores and MOSs, which include Spearman Rank Correlation Coefficient (SRCC), Kendall’s Rank Correlation Coefficient (KRCC), Pearson Linear Correlation Coefficient (PLCC), Root Mean Squared Error (RMSE). 
An excellent quality assessment model should obtain values of SRCC, KRCC, PLCC close to 1 and RMSE to 0.

\begin{table}[tb]\small
\renewcommand\tabcolsep{2.8pt}
    \centering
     
    \begin{tabular}{c|cc|cc|cc} 
    \toprule
    \multirow{2}{*}{Modal}  & \multicolumn{2}{c|}{SJTU-PCQA} & \multicolumn{2}{c|}{WPC} & \multicolumn{2}{c}{WPC2.0} \\ \cline{2-7}
            & SRCC$\uparrow$    & PLCC$\uparrow$   & SRCC$\uparrow$   & PLCC$\uparrow$ & SRCC$\uparrow$   & PLCC$\uparrow$ \\ \hline
        P         &0.8460 &0.8949 & 0.5234 & 0.5552 & 0.5228 & 0.4682\\
        I         &0.8741 &0.8887 & 0.7845 & 0.8084 & 0.7631 & 0.7482\\ 
        P+I       &\bf\textcolor{blue}{0.8786} &\bf\textcolor{blue}{0.8951} & \bf\textcolor{blue}{0.8017} & \bf\textcolor{blue}{0.8137} &\bf\textcolor{blue}{0.7731} & \bf\textcolor{blue}{0.7782} \\
        P+I+SCMA   & \bf\textcolor{red}{0.9103}   & \bf\textcolor{red}{0.9226}    &\bf\textcolor{red}{0.8414}    & \bf\textcolor{red}{0.8556}     &\bf\textcolor{red}{0.8023}    & \bf\textcolor{red}{0.8024}     \\     
    \bottomrule
    \end{tabular}
   \caption{Contributions of the modalities, where `P' stands for only using point cloud features, `I' stands for using only image features, `P+I' represents using both modalities' features through simple concatenation, and `SCMA' indicates the symmetric cross-modality attention is used. }
    \label{tab:modality}
    \vspace{-0.2cm}
\end{table}


\subsubsection{Performance Discussion}
The experimental results are listed in Table \ref{tab:experiment}, from which we can make several useful inspections: a) Our method MM-PCQA presents the best performance among all 3 databases and outperforms the compared NR-PCQA methods by a large margin. For example, MM-PCQA surpasses the second place NR-PCQA method by about 0.05 (0.91 vs. 0.86 (ResSCNN)) on the SJTU-PCQA database, by 0.14 (0.84 vs. 0.70 (PQA-net)) on the WPC database, and by 0.05 (0.80 vs. 0.75 (ResSCNN )) on the WPC2.0 database in terms of SRCC. This is because MM-PCQA enforces the model to relate the compositional quality-aware patterns and mobilize under-complementary information between the image and point cloud modalities to optimize the quality representation by jointly utilizing intra-modal and cross-modal features; b) There are significant performance drops from the SJTU-PCQA database to the WPC and WPC2.0 databases since the WPC and WPC2.0 databases contain more complex distortion settings, which are more challenging for PCQA models. MM-PCQA achieves a relatively smaller performance drop than most compared methods. For example, the SRCC and PLCC values of MM-PCQA drop by 0.07 and 0.08 respectively from the SJTU-PCQA database to the WPC database. However, the top-performing PCQA methods except 3D-NSS experience a larger performance drop over 0.1 on both SRCC and PLCC values. Therefore, we can conclude that MM-PCQA gains more robustness over more complex distortions; c) The image-based handcrafted NR methods BRISQUE, NIQE, and IL-NIQE obtain the poorest performance. This is because such methods are carried out for evaluating the natural scene images, which have a huge gap from the projected images rendered from the point clouds. 
In all, the overall experimental performance  firmly validates our motivation that multi-modal learning can help the model better understand the visual quality of point clouds. 


\subsection{Contributions of the Modalities}
As described above, MM-PCQA jointly employs the features from the point cloud and image modalities and we hold the hypothesis that multi-modal learning helps the model to gain better quality representation than single-modality. Therefore, in this section, we conduct ablation studies to validate our hypothesis. The performance results are presented in Table \ref{tab:modality}. The performance of both single-modal-based model is inferior to the multi-modal-based model, which suggests that both point cloud and image features make contributions to the final results. After using SCMA module, the performance is further boosted, which validates the effectiveness of cross-modality attention.  

\begin{table}[tb]\small

\renewcommand\tabcolsep{2pt}
    \centering
    \begin{tabular}{c|cc|cc|cc}
    \toprule
    \multirow{2}{*}{Model}  & \multicolumn{2}{c|}{SJTU-PCQA} & \multicolumn{2}{c|}{WPC} & \multicolumn{2}{c}{WPC2.0}\\ \cline{2-7}
            & SRCC$\uparrow$    & PLCC$\uparrow$   & SRCC$\uparrow$   & PLCC$\uparrow$ & SRCC$\uparrow$   & PLCC$\uparrow$\\ \hline
        P+FPS         &0.3385 &0.3499 & 0.1226 & 0.1584 & 0.2055 & 0.2744\\
        P+Patch-up    &0.8460 &\bf\textcolor{blue}{0.8949} & 0.5234 & 0.5552 & 0.5228 & 0.4682\\ 
        P+I+FPS       &\bf\textcolor{blue}{0.8512} &{0.8901} & \bf\textcolor{blue}{0.7911} & \bf\textcolor{blue}{0.8122} & \bf\textcolor{blue}{0.7612} & \bf\textcolor{blue}{0.7744}\\
        P+I+Patch-up  & \bf\textcolor{red}{0.9103}   & \bf\textcolor{red}{0.9226}    &\bf\textcolor{red}{0.8414}    & \bf\textcolor{red}{0.8556}     &\bf\textcolor{red}{0.8023}    & \bf\textcolor{red}{0.8024} \\     
    \bottomrule
    \end{tabular}
    \caption{Contributions of the patch-up strategy, where `P' stands for only using point cloud features, `P+I' represents using both point cloud and image features, `FPS' indicates using farthest point sampling strategy, and `Patch-up' indicates using patch-up strategy. }
    \label{tab:patch}
\end{table}

\begin{table}[t]\small
    \centering
    
    \begin{tabular}{c|cc|cc}
    \toprule
    \multirow{2}{*}{Model}  & \multicolumn{2}{c|}{WPC$\rightarrow$SJTU} & \multicolumn{2}{c}{WPC$\rightarrow$WPC2.0} \\ \cline{2-5}
            & SRCC$\uparrow$    & PLCC$\uparrow$   & SRCC$\uparrow$   & PLCC$\uparrow$ \\ \hline
        PQA-net       & \bf\textcolor{blue}{0.5411} & \bf\textcolor{blue}{0.6102} & \bf\textcolor{blue}{0.6006} & \bf\textcolor{blue}{0.6377}\\
        3D-NSS        & 0.1817 & 0.2344 & 0.4933 & 0.5613\\ 
        MM-PCQA       & \bf\textcolor{red}{0.7693}   & \bf\textcolor{red}{0.7779}    &\bf\textcolor{red}{0.7607}    & \bf\textcolor{red}{0.7753}\\     
    \bottomrule
    \end{tabular}
    \caption{Cross-database evaluation, where WPC$\rightarrow$SJTU-PCQA indicates the model is trained on the WPC database and validated with the default testing setup of the SJTU database. Since the WPC and WPC2.0 share some reference point clouds, we remove the groups of point clouds from the WPC database whose references exist in the testing sets of the WPC2.0 database to avoid content overlap.}
    \label{tab:crossdatabase}
    \vspace{-0.2cm}
\end{table}

\begin{table}[t]\small
\renewcommand\tabcolsep{2.2pt}
    \centering
    
    \begin{tabular}{c:c|cc|cc|cc}
    \toprule
    \multirow{2}{24pt}{Type} & \multirow{2}{17pt}{Num}  & \multicolumn{2}{c|}{SJTU-PCQA} & \multicolumn{2}{c|}{WPC} & \multicolumn{2}{c}{WPC2.0}\\ \cline{3-8}
          & & SRCC$\uparrow$    & PLCC$\uparrow$   & SRCC$\uparrow$   & PLCC$\uparrow$ & SRCC$\uparrow$   & PLCC$\uparrow$ \\ \hline
     \multirow{4}{26pt}{sub-models}   
        &2         &0.7421 &0.7824 & 0.4627 & 0.5238 & 0.3446 & 0.4011\\ 
        &4        &0.7317 &0.8214 & 0.4515 & 0.4956 & 0.4270 & 0.4532\\
        &6       &\bf\textcolor{red}{0.8460} &\bf\textcolor{blue}{0.8949} & \bf\textcolor{red}{0.5234} & \bf\textcolor{red}{0.5552} & \bf\textcolor{red}{0.5228} & \bf\textcolor{blue}{0.4682} \\
        &8       &\bf\textcolor{blue}{0.8247} &\bf\textcolor{red}{0.8955} & \bf\textcolor{blue}{0.5024} & \bf\textcolor{blue}{0.5420} &\bf\textcolor{blue}{0.4931}  & \bf\textcolor{red}{0.5232}\\ \hdashline
     \multirow{4}{26pt}{proj-ections}   
        &2       &0.8448 &0.8472 & 0.7417 & 0.7371 & 0.7446 & \bf\textcolor{blue}{0.7587}
        \\ 
        &4       &\bf\textcolor{red}{0.8741} &\bf\textcolor{red}{0.8887} & \bf\textcolor{red}{0.7845} & \bf\textcolor{red}{0.8084} & \bf\textcolor{red}{0.7631} & 0.7482\\
        &6       &0.8601 &\bf\textcolor{blue}{0.8754} & \bf\textcolor{blue}{0.7811} & \bf\textcolor{blue}{0.7976} & \bf\textcolor{blue}{0.7601} & \bf\textcolor{red}{0.7723}\\
        &8       &\bf\textcolor{blue}{0.8612} &0.8577 & 0.7542 & 0.7622 & 0.7521 & 0.7517 \\
    \bottomrule
    \end{tabular}
    \caption{Performance results of the point cloud branch and the image branch by varying the number of sub-models and projections on the SJTU-PCQA, WPC and WPC2.0 databases.}
    \label{tab:ablation_patch}
\end{table}

\begin{table}[!h]\small
    \centering
    \renewcommand\tabcolsep{2.2pt}
    \begin{tabular}{c:c|cc|cc|cc}
    \toprule
    \multirow{2}{*}{P} &\multirow{2}{*}{I} &  \multicolumn{2}{c|}{SJTU-PCQA} & \multicolumn{2}{c|}{WPC} & \multicolumn{2}{c}{WPC2.0} \\ \cline{3-8}
         & & SRCC$\uparrow$    & PLCC$\uparrow$& SRCC$\uparrow$    & PLCC$\uparrow$   & SRCC$\uparrow$   & PLCC$\uparrow$ \\ \hline
        
        DGC & R50         & \bf\textcolor{blue}{0.8734} &\bf\textcolor{blue}{0.8999} & \bf\textcolor{blue}{0.8226} & 0.7984 & \bf\textcolor{blue}{0.7822} & 0.7807 \\
        P++ & R50    & \bf\textcolor{red}{0.9103}   & \bf\textcolor{red}{0.9226}   &\bf\textcolor{red}{0.8414}    & \bf\textcolor{red}{0.8556}    &\bf\textcolor{red}{0.8023}    & \bf\textcolor{red}{0.8024}   \\ \hline
        
        P++ & VGG16         &0.8651 & 0.8669 & 0.7919 & \bf\textcolor{blue}{0.8026} & 0.7744 & \bf\textcolor{blue}{0.7811} \\
        P++ & MNV2   &0.8513 & 0.8721 & 0.8014 & 0.8006 & 0.7614 & 0.7780 \\     
    \bottomrule
    \end{tabular}
    \caption{Performance of different backbones, where `R50' represents the ResNet50 backbone, `MNV2' represents the MobileNetV2 backbone, `DGC' represents the DGCNN backbone, and `P++' represents the PointNet++ backbone respectively.}
    \label{tab:backbone}
\end{table}

\subsection{Ablation for the Patch-up Strategy}
To prove that the patch-up strategy is more suitable for PCQA, we present the performance of using farthest point sampling (FPS) strategy as well. To make the comparison even, 12,288 = 6$\times$2048 points (the same number of points as contained in the default 6 sub-models) are sampled for each point cloud and the results are shown in Table \ref{tab:patch}. It can be seen that the patch-up strategy greatly improves the performance when only point cloud features are used.
The reason is that the sampled points are not able to preserve the local patterns that are vital for quality assessment. Moreover, when the point cloud contains more points, the sampled points may not even maintain the main shape of the object unless greatly increasing the number of sampling points.

\subsection{Cross-database Evaluation}
The cross-database evaluation is further conducted to test the generalization ability of the proposed MM-PCQA and the experimental results are exhibited in Table \ref{tab:crossdatabase}.  Since the WPC database is the largest in scale (740), we mainly train the models on the WPC database and conduct the validation on the SJTU-PCQA (378) and WPC2.0 (400) databases. From the table, we can find that the proposed MM-PCQA better generalizes learned feature representation from the WPC database and achieves much better performance than the most competitive NR-PCQA methods. Furthermore, the WPC$\rightarrow$WPC2.0 MM-PCQA is even superior to all the competitors directly trained on the WPC2.0 database.

\subsection{Number of 2D Projections and 3D sub-models}
We try to investigate the contributions of the point cloud and image branch by varying the number of the 3D sub-models and input 2D projections. The performance results are exhibited in Table \ref{tab:ablation_patch}. We can see that employing 6 sub-models and 4 projections yields the best performance on major aspects for the point cloud and image branch respectively. With the number increasing, the features extracted from sub-models and projections may get redundant, thus resulting in the performance drop.

\subsection{Different Feature Encoders}

In this section, we present the performance of different feature encoders.
The popular 2D image backbones VGG16 \cite{simonyan2014very}, MobileNetV2 \cite{sandler2018mobilenetv2} and ResNet50 \cite{he2016deep} are used for demonstration while the mainstream point cloud backbones PointNet++ \cite{qi2017pointnet++} and DGCNN \cite{wang2019dynamic} are also included. The results are shown in Table \ref{tab:backbone}. It can be found that the proposed PointNet++ and ResNet50 combination is superior to other encoder combinations. Additionally, the performance of different backbones are still competitive compared with other NR-PCQA methods, which further confirms the effectiveness of the proposed framework.


\section{Conclusion}
This paper proposes a novel multi-modal learning approach for no-reference point cloud quality assessment (MM-PCQA). MM-PCQA aims to acquire quality information across modalities and optimize the quality representation. In particular, the point clouds are patched up to preserve the important local geometric structure patterns. Then the projected images are employed to reflect the texture distortions. PointNet++ and ResNet50 are utilized as the feature encoders. Symmetric cross-modal attention is further employed to make full use of the cross-modal information.  Experimental results show that MM-PCQA reaches a new state-of-the-art on the SJTU-PCQA, WPC and WPC2.0 databases. Extensive ablation studies further demonstrate the potency of the proposed multi-modal learning framework.

\subsubsection{Acknowledgements}

This work was supported in part by NSFC (No.62225112, No.61831015), the Fundamental Research Funds for the Central Universities, National Key R\&D Program of China 2021YFE0206700, and Shanghai Municipal Science and Technology Major Project (2021SHZDZX0102).

\bibliographystyle{named}
\bibliography{ijcai23}

\end{document}